\pdfoutput=1
\documentclass[11pt,a4paper]{article}
\usepackage{times,latexsym}
\usepackage{url}
\usepackage[T1]{fontenc}
\usepackage[table]{xcolor}

\usepackage{graphicx}
\usepackage{multirow}
\usepackage{tabularx}
\usepackage{amsmath}
\usepackage{algorithmic}
\usepackage[ruled]{algorithm2e}
\usepackage{tikz}
\usepackage{makecell}
\usepackage{comment}
\usepackage{amssymb}
\usepackage{arydshln}
\usepackage{enumitem}
\usepackage{cancel}
\usepackage{array,multirow,graphicx}
\usepackage{rotating}
\usepackage{pgfplots}
\usepackage{hyperref}
\usepackage[]{acl2020}

\aclfinalcopy

\title{Learning a  Multi-Domain Curriculum for  Neural Machine Translation}

\author{Wei Wang\\
  Google Research \\
  {\tt wangwe@google.com} \\\And
  Ye Tian \\
  Google Research \\
  {\tt ytian@google.com} \\ \And
  Jiquan Ngiam
 \\
  Google Brain \\
  {\tt jngiam@google.com} \\ \AND
  Yinfei Yang
 \\
  Google Research \\
  {\tt yinfeiy@google.com} \\ \And
   Isaac Caswell
 \\
  Google Research \\
  {\tt icaswell@google.com} \\ \And
    Zarana Parekh \\
  Google Research \\
  {\tt zarana@google.com}
  }

\date{}
\begin{document}
\maketitle
\begin{abstract}
Most data selection research in  machine translation focuses on improving a single domain. We perform data selection for multiple domains at once. This is achieved by carefully introducing instance-level domain-relevance features and automatically constructing a training curriculum to gradually concentrate on multi-domain relevant and noise-reduced data batches. Both the choice of features and 
the use of curriculum are crucial for balancing and improving all domains, including out-of-domain.
In large-scale experiments, the multi-domain curriculum simultaneously reaches or outperforms the individual performance and brings solid gains over no-curriculum training.
\end{abstract}

\section{Introduction}
In  machine translation (MT), data selection,  e.g., \cite{Moore2010,Axelrod2011}, has remained as a fundamental and important research topic. It has played a crucial role in domain adaptation by selecting domain-matching training examples, or data cleaning (aka denoising) by selecting high-quality examples. 
So far, the most extensively studied scenario assumes a single domain to improve.

It becomes both technically challenging and practically appealing to build a large-scale multi-domain neural machine translation (NMT) model that performs simultaneously well on multiple domains at once. This requires addressing research challenges such as catastrophic forgetting \cite{Goodfellow2014} at scale and  data balancing. Such a model can easily find potential use cases, i.e., as a solid general service, for downstream transfer learning, for better deployment efficiency, or for transfer learning across datasets.

Unfortunately, existing single-domain data-selection methods do not work well for multiple domains.
For example, improving the translation accuracy of one domain will often hurt that of another \cite{dynamiccds,mixing},
 and improving model generalization across all domains by   clean-data selection \cite{koehn-EtAl:2018:WMT}  may not promise  optimization of  a particular  domain. 
 Multiple aspects need to be considered for training a multi-domain model. 

This paper presents a dynamic data selection method to multi-domain NMT.
Things we do differently from previous work in mixing data are the choice of instance-level features and the employment of a multi-domain curriculum that is additionally able to denoise. These are crucial for mixing and improving all domains, including out-of-domain.
We experiment with large datasets at different noise levels and show that the resulting models meet our requirements.

\section{Related Work}\label{related_work}

\begin{table}[t]
\begin{center}
\small
\begin{tabular}{*{4}{|c}|}
\cline{3-4}
\multicolumn{2}{c}{} & \multicolumn{1}{|c|}{{\bf Static}} &  \multicolumn{1}{|c|}{{\bf Dynamic}} \\
\hline
\multirow{2}{3.5em}{{\bf Single}} & domain & Y & Y \\
& noise & Y & Y \\ 
\hline
\multirow{2}{3.5em}{{\bf Multi}} & domain & Y & \cellcolor{gray!15}  \\ 
& noise & N &  \cellcolor{gray!15} \multirow{-2}{6em}{N {\small (Our Work)}}  \\ 
\hline
\end{tabular}

\end{center}
\vspace{-0.1in}
\caption{\footnotesize Data selection and data mixing research in NMT. `Y': There is previous research that studies this case. `N': No previous research has studied this case.  \label{landscape}}
\end{table}

In MT, research that is most relevant to our work is data selection and data mixing, both being concerned with how to sample examples to train an MT model, usually for domain adaptation.  Table~\ref{landscape} categorizes previous research by two aspects and shows where our work stands. These two aspects are:
\begin{enumerate}
    \item Is the method concerned with a single domain or multiple domains?
    \item Does the method use data statically or dynamically?
\end{enumerate}


\paragraph{Static data selection for a single domain.} \citet{Moore2010} select  in-domain data  for n-gram language model (LM) training.  It is later generalized by \citet{Axelrod2011} to select parallel data for training MT models. 
\citet{chen2016a,chen2016b} use classifiers to select domain data. Clean-data selection \cite{koehn-filter:2019:WMT,koehn-EtAl:2018:WMT,junczysdowmunt:2018:WMT2}  reduces harmful data noise to improve translation quality across domains. All these works select a data subset for a single ``domain''\footnote{We treat denoising as a domain in the paper, inspired by previous works that treat data noise using domain adaptation methods, e.g., \citep{junczysdowmunt:2018:WMT2}.} and treat the selected data as a static/flat distribution.

\paragraph{Dynamic data selection for a single domain.}
 Static selection has two shortcomings: it discards data and it treats all examples equally after selection. When data is scarce, any data could be helpful, even if it is out of domain or noisy\footnote{
We refer to data regularization (using more data) and to transfer learning (fine-tuning) to exploit both data quantity  and quality, the idea behind dynamic data selection. See \ref{noisy_exp}.}. Dynamic data selection  is introduced to ``sort'' data from least in-domain  to most in-domain. Training NMT models on data sorted this way effectively takes advantage of transfer learning. Curriculum learning (CL) \cite{curri_learn} has been used as a formulation for dynamic data selection.
Domain curricula \cite{dynamiccds,curriculum_domain} are used for domain adaptation.  Model stacking \cite{model_stacking,DBLP:journals/corr/FreitagA16} is a practical idea to build domain models.
CL is also used for denoising \cite{curriculum_opt_for_nmt,wang-etal-2018-dynamic,denoise_nmt}, and for faster  convergence and improved general quality \cite{DBLP:journals/corr/abs-1811-00739,DBLP:journals/corr/abs-1903-09848}.
\citet{wang-etal-2018-dynamic} introduce a curriculum for training efficiency.
In addition to data sorting/curriculum, instance/loss weighting \citep{D17-1155,chen-etal-2017-cost,dds} has been used as an alternative. 
CL for NMT represents the SOTA data-selection method, but most existing works target at  a single ``domain'', be it a specific domain or the ``denoising domain''.

\paragraph{Static data mixing for multiple domains.} When mixing data from multiple domains, a fundamental challenge is to address catastrophic forgetting \cite{Goodfellow2014}--training an NMT model to focus on one domain can likely hurt another \cite{dynamiccds,mixing}. \citet{mixing} learn domain-discerning (or -invariant) network representation with a domain discriminator network for NMT.   The methods, however, require that domain labels are available in data. \citet{DBLP:journals/corr/abs-1805-02282} cluster data and tag each cluster as multi-domain NMT training data, but the method treats data in each cluster as a flat distribution. 
\citet{farajian-etal-2017-multi} implement multi-domain NMT by
on-the-fly data retrieval and adaptation per sentence, at increased inference cost. 
Most existing methods (or experiment setups) have the following problems: (i) They mix data statically. (ii) They don't consider the impact of data noise, which is a source of catastrophic forgetting. (iii) Experiments are carried out with small datasets, without separate examination on the data regularization effect. (iv) They do not examine  out-of-domain performamce. 
 
\paragraph{Automatic data balancing for multi-domains.}
\citep{xinyi:2020:acl} automatically learn to weight (flat) data streams of multi-languages (or "domains"). We perform dynamic data selection and regularization through a mulit-domain curriculum.  










\paragraph{Automatic curriculum learning.} Our work falls under automatic curriculum construction \cite{DBLP:journals/corr/GravesBMMK17} and is directly inspired by \citet{tsvetkov-etal-2016-learning}, who
learn to weight and combine instance-level features to form a curriculum for an embedding learning task, through Bayesian Optimization. A similar idea \cite{DBLP:journals/corr/RuderP17} is  used to improve other NLP tasks.  Here, we use the idea for NMT to  construct a multi-domain data selection scheme with various selection scores at our disposal. The problem we study is connected to the more general multi-objective optimization problem. 
\citet{multi-objective} uses Bandit learning to tune hyper-parameters such as the number of network layers for NMT.

\paragraph{More related work.} 
Previously, catastrophic forgetting has mostly been studied in the continued-training setup \citep{saunders-etal-2019-domain, khayrallah-etal-2018-regularized}, to refer to the degrading performance on the out-of-domain task when a model is fine-tuned on in-domain data. This setup is a popular topic in general machine learning research \cite{NIPS2019_9354}. \citet{thompson-etal-2018-freezing} study domain adaptation by freezing subnetworks. Our work instead addresses forgetting in the data-balancing scenario for multi-domains. We use curriculum to generalize fine-tuning.

\section{Curriculum Learning for NMT}\label{sec:CL}
We first introduce curriculum learning (CL) \cite{curri_learn}, which serves as a formulation for SOTA single-domain dynamic data selection and which our method is built upon and generalizes. In CL, a curriculum, $\mathcal{C}$,
is a sequence of training criteria over training steps.
 A training criterion, $Q_t(y|x)$, at step $t$ is associated with a set of weights,
 $W_t(x,y)$,\footnote{
 As a preview, in our paper, $W_t(x,y)$ uses uniform weights  over selected examples and assigns zero weights for filtered examples, similar to a mask.
 }
 over training sentence pairs $(x,y)$ in a parallel dataset $D$, where
 $y$ is the translation for $x$. $Q_t(y|x)$ is a re-weighting of the original training distribution $P(y|x)$:
\begin{equation} 
\begin{split}
Q_t(y|x) \propto W_t(x,y)P(y|x), 
  \hspace{0.1in} \forall (x,y) \in D \label{form1}
\end{split}
\end{equation}
Hence, for $T$ maximum training steps, $\mathcal{C}$ is a sequence:
\begin{eqnarray} 
  \mathcal{C} = \langle Q_1, ..., Q_t, ..., Q_T \rangle  \label{form2}
\end{eqnarray}
At step $t$, an online learner randomly samples a data batch from $Q_t$ to fine-tune model $m_{t-1}$ into $m_t$. 
Therefore,  $\mathcal{C}$ corresponds to a sequence of models,
\begin{eqnarray}
\langle m_1, ..., m_t, ..., M \rangle.
\end{eqnarray}
$M$ is the
 final model that the entire curriculum has been optimizing towards.
  Intermediate models,  $m_t$, serve as ``stepping stones'' to $M$, to transfer knowledge through them and regularize the training for generalization.
   A performance metric
 $\mathcal{P}(\mathcal{C})$ evaluates $M$ on a development or test set, after training on $\mathcal{C}$.


\begin{center}
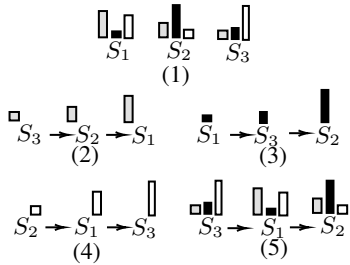
\begin{figure}[t]
\small

\begin{center}

\tikzset{every picture/.style={line width=0.75pt}} 

\begin{tikzpicture}[x=0.4pt,y=0.4pt,yscale=-0.8,xscale=0.8]


\draw  [fill={rgb, 255:red, 224; green, 224; blue, 224 }  ,fill opacity=1 ] (126,5.92) -- (135.5,5.92) -- (135.5,36.92) -- (126,36.92) -- cycle ;
\draw  [fill={rgb, 255:red, 0; green, 0; blue, 0 }  ,fill opacity=1 ] (141.5,29.92) -- (151.5,29.92) -- (151.5,36.92) -- (141.5,36.92) -- cycle ;
\draw  [fill={rgb, 255:red, 255; green, 255; blue, 255 }  ,fill opacity=1 ](156,10.92) -- (165.5,10.92) -- (165.5,37.92) -- (156,37.92) -- cycle ;
\draw  [fill={rgb, 255:red, 224; green, 224; blue, 224 }  ,fill opacity=1 ] (196.5,19.92) -- (206.5,19.92) -- (206.5,36.92) -- (196.5,36.92) -- cycle ;
\draw  [fill={rgb, 255:red, 0; green, 0; blue, 0 }  ,fill opacity=1 ] (212.5,-1.08) -- (220.5,-1.08) -- (220.5,36.92) -- (212.5,36.92) -- cycle ;
\draw  [fill={rgb, 255:red, 255; green, 255; blue, 255 }  ,fill opacity=1 ](225.5,27.92) -- (236.5,27.92) -- (236.5,37.92) -- (225.5,37.92) -- cycle ;
\draw  [fill={rgb, 255:red, 224; green, 224; blue, 224 }  ,fill opacity=1 ] (265.5,28.92) -- (276.5,28.92) -- (276.5,38.92) -- (265.5,38.92) -- cycle ;
\draw  [fill={rgb, 255:red, 255; green, 255; blue, 255 }  ,fill opacity=1 ] (294.5,-0.08) -- (301.5,-0.08) -- (301.5,39.92) -- (294.5,39.92) -- cycle ;
\draw  [fill={rgb, 255:red, 0; green, 0; blue, 0 }  ,fill opacity=1 ] (281.5,25.92) -- (289.5,25.92) -- (289.5,38.92) -- (281.5,38.92) -- cycle ;
\draw  [fill={rgb, 255:red, 224; green, 224; blue, 224 }  ,fill opacity=1 ] (21.5,124.92) -- (32.5,124.92) -- (32.5,134.92) -- (21.5,134.92) -- cycle ;
\draw  [fill={rgb, 255:red, 224; green, 224; blue, 224 }  ,fill opacity=1 ] (89.5,118.92) -- (99.5,118.92) -- (99.5,135.92) -- (89.5,135.92) -- cycle ;
\draw  [fill={rgb, 255:red, 224; green, 224; blue, 224 }  ,fill opacity=1 ] (156,105.92) -- (165.5,105.92) -- (165.5,136.92) -- (156,136.92) -- cycle ;
\draw    (68.5,151.92) -- (88.5,151.92) ;
\draw [shift={(90.5,151.92)}, rotate = 180] [color={rgb, 255:red, 0; green, 0; blue, 0 }  ][line width=0.75]    (10.93,-3.29) .. controls (6.95,-1.4) and (3.31,-0.3) .. (0,0) .. controls (3.31,0.3) and (6.95,1.4) .. (10.93,3.29)   ;

\draw    (134.5,151.92) -- (154.5,151.92) ;
\draw [shift={(156.5,151.92)}, rotate = 180] [color={rgb, 255:red, 0; green, 0; blue, 0 }  ][line width=0.75]    (10.93,-3.29) .. controls (6.95,-1.4) and (3.31,-0.3) .. (0,0) .. controls (3.31,0.3) and (6.95,1.4) .. (10.93,3.29)   ;

\draw  [fill={rgb, 255:red, 0; green, 0; blue, 0 }  ,fill opacity=1 ] (247.5,128.92) -- (257.5,128.92) -- (257.5,135.92) -- (247.5,135.92) -- cycle ;
\draw  [fill={rgb, 255:red, 0; green, 0; blue, 0 }  ,fill opacity=1 ] (314.5,124.92) -- (322.5,124.92) -- (322.5,137.92) -- (314.5,137.92) -- cycle ;
\draw  [fill={rgb, 255:red, 0; green, 0; blue, 0 }  ,fill opacity=1 ] (386.5,98.92) -- (394.5,98.92) -- (394.5,136.92) -- (386.5,136.92) -- cycle ;
\draw    (279.5,151.92) -- (299.5,151.92) ;
\draw [shift={(301.5,151.92)}, rotate = 180] [color={rgb, 255:red, 0; green, 0; blue, 0 }  ][line width=0.75]    (10.93,-3.29) .. controls (6.95,-1.4) and (3.31,-0.3) .. (0,0) .. controls (3.31,0.3) and (6.95,1.4) .. (10.93,3.29)   ;

\draw    (348.5,149.92) -- (368.5,149.92) ;
\draw [shift={(370.5,149.92)}, rotate = 180] [color={rgb, 255:red, 0; green, 0; blue, 0 }  ][line width=0.75]    (10.93,-3.29) .. controls (6.95,-1.4) and (3.31,-0.3) .. (0,0) .. controls (3.31,0.3) and (6.95,1.4) .. (10.93,3.29)   ;

\draw  [fill={rgb, 255:red, 224; green, 224; blue, 224 }  ,fill opacity=1 ] (376.5,226.92) -- (386.5,226.92) -- (386.5,243.92) -- (376.5,243.92) -- cycle ;
\draw  [fill={rgb, 255:red, 0; green, 0; blue, 0 }  ,fill opacity=1 ] (392.5,205.92) -- (400.5,205.92) -- (400.5,243.92) -- (392.5,243.92) -- cycle ;
\draw  [fill={rgb, 255:red, 255; green, 255; blue, 255 }  ,fill opacity=1 ] (405.5,234.92) -- (416.5,234.92) -- (416.5,244.92) -- (405.5,244.92) -- cycle ;
\draw  [fill={rgb, 255:red, 224; green, 224; blue, 224 }  ,fill opacity=1 ] (307,214.92) -- (316.5,214.92) -- (316.5,245.92) -- (307,245.92) -- cycle ;
\draw  [fill={rgb, 255:red, 0; green, 0; blue, 0 }  ,fill opacity=1 ] (322.5,238.92) -- (332.5,238.92) -- (332.5,245.92) -- (322.5,245.92) -- cycle ;
\draw  [fill={rgb, 255:red, 255; green, 255; blue, 255 }  ,fill opacity=1 ](337,219.92) -- (346.5,219.92) -- (346.5,246.92) -- (337,246.92) -- cycle ;
\draw  [fill={rgb, 255:red, 224; green, 224; blue, 224 }  ,fill opacity=1 ] (233.5,234.92) -- (244.5,234.92) -- (244.5,244.92) -- (233.5,244.92) -- cycle ;
\draw  [fill={rgb, 255:red, 255; green, 255; blue, 255 }  ,fill opacity=1 ] (262.5,205.92) -- (269.5,205.92) -- (269.5,245.92) -- (262.5,245.92) -- cycle ;
\draw  [fill={rgb, 255:red, 0; green, 0; blue, 0 }  ,fill opacity=1 ] (249.5,231.92) -- (257.5,231.92) -- (257.5,244.92) -- (249.5,244.92) -- cycle ;
\draw    (276.5,261.92) -- (296.5,261.92) ;
\draw [shift={(298.5,261.92)}, rotate = 180] [color={rgb, 255:red, 0; green, 0; blue, 0 }  ][line width=0.75]    (10.93,-3.29) .. controls (6.95,-1.4) and (3.31,-0.3) .. (0,0) .. controls (3.31,0.3) and (6.95,1.4) .. (10.93,3.29)   ;

\draw    (349.5,260.92) -- (369.5,260.92) ;
\draw [shift={(371.5,260.92)}, rotate = 180] [color={rgb, 255:red, 0; green, 0; blue, 0 }  ][line width=0.75]    (10.93,-3.29) .. controls (6.95,-1.4) and (3.31,-0.3) .. (0,0) .. controls (3.31,0.3) and (6.95,1.4) .. (10.93,3.29)   ;

\draw  [fill={rgb, 255:red, 255; green, 255; blue, 255 }  ,fill opacity=1 ] (46.5,235.92) -- (57.5,235.92) -- (57.5,245.92) -- (46.5,245.92) -- cycle ;
\draw  [fill={rgb, 255:red, 255; green, 255; blue, 255 }  ,fill opacity=1 ] (119,218.92) -- (128.5,218.92) -- (128.5,245.92) -- (119,245.92) -- cycle ;
\draw  [fill={rgb, 255:red, 255; green, 255; blue, 255 }  ,fill opacity=1 ](184.5,206.92) -- (191.5,206.92) -- (191.5,246.92) -- (184.5,246.92) -- cycle ;
\draw    (63.5,260.92) -- (83.5,260.92) ;
\draw [shift={(85.5,260.92)}, rotate = 180] [color={rgb, 255:red, 0; green, 0; blue, 0 }  ][line width=0.75]    (10.93,-3.29) .. controls (6.95,-1.4) and (3.31,-0.3) .. (0,0) .. controls (3.31,0.3) and (6.95,1.4) .. (10.93,3.29)   ;

\draw    (135.5,260.92) -- (155.5,260.92) ;
\draw [shift={(157.5,260.92)}, rotate = 180] [color={rgb, 255:red, 0; green, 0; blue, 0 }  ][line width=0.75]    (10.93,-3.29) .. controls (6.95,-1.4) and (3.31,-0.3) .. (0,0) .. controls (3.31,0.3) and (6.95,1.4) .. (10.93,3.29)   ;


\draw (149,53) node   {$S_{1}$};

\draw (220,53) node   {$S_{2}$};
\draw (289,55) node   {$S_{3}$};
\draw (45,151) node   {$S_{3}$};

\draw (113,152) node   {$S_{2}$};
\draw (179,153) node   {$S_{1}$};
\draw (255,152) node   {$S_{1}$};
\draw (322,154) node   {$S_{3}$};
\draw (394,153) node   {$S_{2}$};
\draw (400,260) node   {$S_{2}$};
\draw (330,262) node   {$S_{1}$};
\draw (257,261) node   {$S_{3}$};

\draw (216,82) node  [align=left] {(1)};
\draw (110,177) node  [align=left] {(2)};
\draw (330,178) node  [align=left] {(3)};
\draw (331,287) node  [align=left] {(5)};
\draw (41,261) node   {$S_{2}$};
\draw (112,261) node   {$S_{1}$};
\draw (179,262) node   {$S_{3}$};
\draw (112,291) node  [align=left] {(4)};

\end{tikzpicture}

\end{center}
\vspace{-0.1in}
\caption{\small{Data order in single-domain curricula and a potential multi-domain curriculum. (1) A toy training dataset of 3 examples. Each example has three scores, representing relevance to three domains, grey/dark/white domains, respectively. The higher the bar the more relevant. (2) Grey-domain order. (3) Dark-domain order. (4) White-domain order. (5) A potential multi-domain data order.
}}
\label{curriculum}

\end{figure}
\end{center}

In NMT, CL is used to implement dynamic data selection. First, a scoring function (Section~\ref{features}) is employed to measure the usefulness of an example to a domain and sort data.
 Then mini-batch sampling, e.g., \cite{R17-1050}, is  designed to realize the weighting $W_t$, to dynamically evolve the training criteria $Q_t$ towards in-domain. 
Figure~\ref{curriculum} (1)-(4) illustrates the basic idea of the curriculum we use. (1) shows three sentence pairs, $S_1, S_2, S_3$, each having three scores,  respectively representing usefulness to three domains. A grey-domain training curriculum, for example, relies on the data order in (2), gradually discards least useful examples according to $W_t(x,y)$ (Eq.~\ref{form1})  in Table~\ref{curriculum_example} (1): At step 1, the learner uniformly samples from all examples ($W_1$), producing model $m_1$. In step 2,  the least-in-domain $S_3$ is discarded (strikethrough) by $W_2$ so we sample from subset $\{S_1, S_2\}$ uniformly to reach $m_2$. We repeat this until reaching the final model $M$. In this process, sampling is uniform in each step, but in-domain examples (e.g., $S_1$) are reused more over steps. Similarly, we can construct the dark-domain curriculum in Figure~\ref{curriculum} (3) and the white-domain (4).


\begin{table}[t]
\small
\centering
\begin{tabularx}{\textwidth}{@{}llX@{}}
$\hspace{0.05in}$ $W_1$ $\hspace{0.03in}\to\hspace{0.03in}$   $W_2$ $\hspace{0.03in}\to\hspace{0.03in}$  $W_3$  
$\hspace{0.5in}$ $W_1$ $\hspace{0.03in}\to\hspace{0.03in}$   $W_2$ $\hspace{0.03in}\to\hspace{0.03in}$  $W_3$\\ 
  $\begin{pmatrix}1/3\\1/3\\1/3\end{pmatrix}$
  $\begin{pmatrix}1/2 \\ 1/2 \\ \cancel{0.0} \end{pmatrix}$ 
  $\begin{pmatrix}1.0\\\cancel{0.0}\\\cancel{0.0}\end{pmatrix}$
  $\hspace{0.2in}$ 
  $\begin{pmatrix}1/3\\1/3\\1/3\end{pmatrix}$
  $\begin{pmatrix}1/2 \\ 1/2 \\ \cancel{0.0} \end{pmatrix}$ 
  $\begin{pmatrix}\cancel{0.0}\\1.0\\\cancel{0.0}\end{pmatrix}$ \\
  $\hspace{0.6in}$ (1) $\hspace{1.3in}$ (2) \\
\end{tabularx}
\vspace{-0.1in}
\caption{\small Curriculum examples characterized by re-weighting, $W_t(x, y)$, 
over three steps, to stochastically order data to benefit a final domain.  Strikethrough discards examples. (1) corresponds to data order  Figure~\ref{curriculum} (2). (2) corresponds to data order  Figure~\ref{curriculum} (5).
 \label{curriculum_example}}
\end{table}



\section{Our Approach: Learning a Multi-Domain Curriculum}
\subsection{General Idea} \label{general_idea}

The challenges in multi-domain/-task data selection lie in addressing catastrophic forgetting and data balancing. In Figure~\ref{curriculum}, while curriculum (2) moves a model to the grey-domain direction, this direction may not necessarily be positively consistent with the dark domain (Figure~\ref{curriculum} (3)), causing dropped dark-domain performance. Ideally, a training example that introduces the least forgetting across all domains would have gradients that move the model in a common direction towards all domains. While this may not be easily feasible by selecting a single example, we would like the intuition to work in a data batch on average. Therefore, our idea is to carefully introduce per-example data-selection scores (called features) to measure ``domain sharing'', intelligently weight them  to balance the domains of interest, and dynamically schedule examples  to trade-off between regularization and domain adaptation.  

A method to realize the above idea has the following {\em properties}:
\begin{enumerate}
    \item Features of an example reflect its relevance to domains. 
    \item Feature weights are jointly learned/optimized based on end model performance.
    \item Training is dynamic, by gradually focusing on multi-domain relevant and noise-reduced data batches.
\end{enumerate}
Furthermore, a viable multi-domain curriculum  meets the following {\em performance requirements}: 
\begin{itemize}
\item[(i)] It improves the baseline model across all domains.
\item[(ii)] It simultaneously reaches (or outperforms) the peak performance of individual single-domain curricula. 
\end{itemize}
Above  requires  improvement over out-of-domain, too.




\subsection{The Framework}


\begin{figure}[t]
\small


\tikzset{every picture/.style={line width=0.75pt}} 

\begin{tikzpicture}[x=0.75pt,y=0.75pt,yscale=-0.65,xscale=0.6]

\draw   (197.5,156) .. controls (197.5,135.01) and (227.72,118) .. (265,118) .. controls (302.28,118) and (332.5,135.01) .. (332.5,156) .. controls (332.5,176.99) and (302.28,194) .. (265,194) .. controls (227.72,194) and (197.5,176.99) .. (197.5,156) -- cycle ;
\draw    (154.5,163) -- (188.54,156.38) ;
\draw [shift={(190.5,156)}, rotate = 529] [color={rgb, 255:red, 0; green, 0; blue, 0 }  ][line width=0.75]    (10.93,-3.29) .. controls (6.95,-1.4) and (3.31,-0.3) .. (0,0) .. controls (3.31,0.3) and (6.95,1.4) .. (10.93,3.29)   ;

\draw   (349.5,110.31) .. controls (349.5,93.02) and (374.12,79) .. (404.5,79) .. controls (434.88,79) and (459.5,93.02) .. (459.5,110.31) .. controls (459.5,127.61) and (434.88,141.63) .. (404.5,141.63) .. controls (374.12,141.63) and (349.5,127.61) .. (349.5,110.31) -- cycle ;
\draw    (159.5,51) -- (182.5,51) ;
\draw [shift={(184.5,51)}, rotate = 180] [color={rgb, 255:red, 0; green, 0; blue, 0 }  ][line width=0.75]    (10.93,-3.29) .. controls (6.95,-1.4) and (3.31,-0.3) .. (0,0) .. controls (3.31,0.3) and (6.95,1.4) .. (10.93,3.29)   ;

\draw    (159.5,92) -- (183.74,78.95) ;
\draw [shift={(185.5,78)}, rotate = 511.7] [color={rgb, 255:red, 0; green, 0; blue, 0 }  ][line width=0.75]    (10.93,-3.29) .. controls (6.95,-1.4) and (3.31,-0.3) .. (0,0) .. controls (3.31,0.3) and (6.95,1.4) .. (10.93,3.29)   ;

\draw  [dash pattern={on 4.5pt off 4.5pt}]  (147.5,75) -- (182.58,64.57) ;
\draw [shift={(184.5,64)}, rotate = 523.44] [color={rgb, 255:red, 0; green, 0; blue, 0 }  ][line width=0.75]    (10.93,-3.29) .. controls (6.95,-1.4) and (3.31,-0.3) .. (0,0) .. controls (3.31,0.3) and (6.95,1.4) .. (10.93,3.29)   ;

\draw    (330.5,145) -- (349.42,115.68) ;
\draw [shift={(350.5,114)}, rotate = 482.83] [color={rgb, 255:red, 0; green, 0; blue, 0 }  ][line width=0.75]    (10.93,-3.29) .. controls (6.95,-1.4) and (3.31,-0.3) .. (0,0) .. controls (3.31,0.3) and (6.95,1.4) .. (10.93,3.29)   ;

\draw    (270.5,84) -- (259.26,111.15) ;
\draw [shift={(258.5,113)}, rotate = 292.48] [color={rgb, 255:red, 0; green, 0; blue, 0 }  ][line width=0.75]    (10.93,-3.29) .. controls (6.95,-1.4) and (3.31,-0.3) .. (0,0) .. controls (3.31,0.3) and (6.95,1.4) .. (10.93,3.29)   ;

\draw    (409.5,145) -- (408.58,168) ;
\draw [shift={(408.5,170)}, rotate = 272.29] [color={rgb, 255:red, 0; green, 0; blue, 0 }  ][line width=0.75]    (10.93,-3.29) .. controls (6.95,-1.4) and (3.31,-0.3) .. (0,0) .. controls (3.31,0.3) and (6.95,1.4) .. (10.93,3.29)   ;

\draw  [dash pattern={on 4.5pt off 4.5pt}] (87.5,6) -- (477.5,6) -- (477.5,206) -- (87.5,206) -- cycle ;
\draw  [dash pattern={on 4.5pt off 4.5pt}]  (70.5,47) -- (88.5,47) ;
\draw [shift={(90.5,47)}, rotate = 180] [color={rgb, 255:red, 0; green, 0; blue, 0 }  ][line width=0.75]    (10.93,-3.29) .. controls (6.95,-1.4) and (3.31,-0.3) .. (0,0) .. controls (3.31,0.3) and (6.95,1.4) .. (10.93,3.29)   ;

\draw  [dash pattern={on 4.5pt off 4.5pt}]  (69.5,76) -- (90.5,75.09) ;
\draw [shift={(92.5,75)}, rotate = 537.51] [color={rgb, 255:red, 0; green, 0; blue, 0 }  ][line width=0.75]    (10.93,-3.29) .. controls (6.95,-1.4) and (3.31,-0.3) .. (0,0) .. controls (3.31,0.3) and (6.95,1.4) .. (10.93,3.29)   ;

\draw  [dash pattern={on 4.5pt off 4.5pt}]  (73.5,102) -- (90.5,102) ;
\draw [shift={(92.5,102)}, rotate = 180] [color={rgb, 255:red, 0; green, 0; blue, 0 }  ][line width=0.75]    (10.93,-3.29) .. controls (6.95,-1.4) and (3.31,-0.3) .. (0,0) .. controls (3.31,0.3) and (6.95,1.4) .. (10.93,3.29)   ;

\draw  [dash pattern={on 4.5pt off 4.5pt}]  (410.5,194) -- (410.5,228) ;
\draw [shift={(410.5,230)}, rotate = 270] [color={rgb, 255:red, 0; green, 0; blue, 0 }  ][line width=0.75]    (10.93,-3.29) .. controls (6.95,-1.4) and (3.31,-0.3) .. (0,0) .. controls (3.31,0.3) and (6.95,1.4) .. (10.93,3.29)   ;

\draw   (212.5,240) .. controls (212.5,228.95) and (232.09,220) .. (256.25,220) .. controls (280.41,220) and (300,228.95) .. (300,240) .. controls (300,251.05) and (280.41,260) .. (256.25,260) .. controls (232.09,260) and (212.5,251.05) .. (212.5,240) -- cycle ;
\draw    (329.5,241) -- (306.5,241) ;
\draw [shift={(304.5,241)}, rotate = 360] [color={rgb, 255:red, 0; green, 0; blue, 0 }  ][line width=0.75]    (10.93,-3.29) .. controls (6.95,-1.4) and (3.31,-0.3) .. (0,0) .. controls (3.31,0.3) and (6.95,1.4) .. (10.93,3.29)   ;

\draw    (62.5,240) -- (211.5,240) ;

\draw    (61.5,240) -- (61.5,123) ;
\draw [shift={(61.5,121)}, rotate = 450] [color={rgb, 255:red, 0; green, 0; blue, 0 }  ][line width=0.75]    (10.93,-3.29) .. controls (6.95,-1.4) and (3.31,-0.3) .. (0,0) .. controls (3.31,0.3) and (6.95,1.4) .. (10.93,3.29)   ;

\draw (129,166) node   {$D$};
\draw (265,146) node  [align=left]
{curriculum};
\draw (405.5,112.31) node  [align=center] {finetune \\ NMT};
\draw (128,47) node   {$f_{1}( x,y)$};
\draw (131,104) node   {$f_{N}( x,y)$};
\draw (111,74) node   {$\vdots $};
\draw (415.5,181.31) node  [align=left] {model};
\draw (58,44) node   {$v_{1}$};
\draw (59,100) node   {$v_{N}$};
\draw (58,74) node   {$\vdots $};
\draw (432,243) node   {$\mathcal{P}_{1}$};
\draw (477,243) node   {$\mathcal{P}_{K}$};
\draw (451,241) node   {$\dotsc $};
\draw (377,242) node  [align=left] {\underline{\textit{eval:}}};
\draw (257,239) node  [align=left] {optimizer};
\draw (320,50) node   {$f( x,y)=V \cdotp F( x,y)$};
\draw (273,174) node   {${\mathcal{\widehat{C}}}(V)$};
\draw (34,70) node   {$V=$};

\end{tikzpicture}

\vspace{-0.1in}
\caption {\footnotesize Learning a multi-domain curriculum.}
\label{multitask_curriculum}

\end{figure}
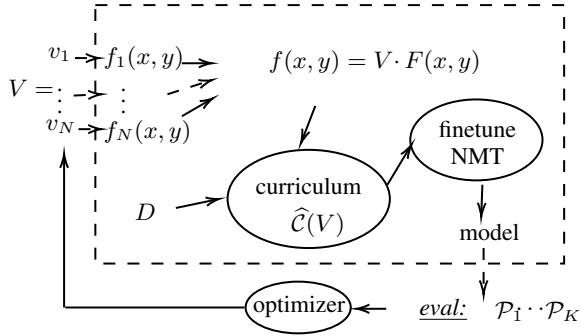

Formally, for a sentence pair $(x, y)$,  let $f_n(x, y) \in \mathbb{R}$
be its $n$-th feature that specifies how $(x, y)$ is useful to a domain. Suppose we are interested in $K$ domains and each example has $N$ features. For instance, each sentence pair of $S1, S2, S3$ in Figure~\ref{curriculum} (1) has three features ($N=3$), each for one domain ($K=3$).\footnote{
But $N$ does not necessarily equal $K$ because we can introduce multiple features for one domain or a single feature for multiple domains.} We represent $(x,y)$'s 
features using a feature vector
$F(x,y)=[f_0(x, y), ..., f_{N-1}(x, y)]$.
Given a weight vector $V=\left[v_0, ..., v_{N-1}\right]$ for all sentence pairs, we compute an aggregated score
\begin{eqnarray}
f(x, y) = V \cdot F(x,y) \label{varphi2}
\end{eqnarray}
for each sentence pair and sort the entire data in increasing order.
We then construct a curriculum $\mathcal{\widehat{C}}(V)$ to {\em fine-tune} a warmed-up model, evaluate its performance and propose a next weight vector. After several iterations/trials, the optimal weight vector  $V^{*}$ is the one with the best end performance:
\begin{eqnarray}
V^* = \arg\max_{V} \mathcal{P}(\widehat{\mathcal{C}}(V)) \label{W}
\end{eqnarray}
Figure~\ref{multitask_curriculum} shows the framework.
For the process to be practical and scalable, $\widehat{\mathcal{C}}$ fine-tunes a warmed-up model for  a small number of steps. The learned $V^*$ can then eventually be used for retraining  a final model from scratch.

\subsection{Instance-Level Features} \label{features}
We design the following types of features for each training example and instantiate them in Experiments (Section~\ref{sec:exp}).

\paragraph{NMT domain features ($q_Z$)} compute, for a pair $(x, y)$, the
cross-entropy difference between two NMT models:
\begin{eqnarray}
q_Z\left(x,y\right) \hspace{-0.04in} = \hspace{-0.05in}
 \frac{\log P\left(y|x; \theta_Z\right) \hspace{-0.04in} - \hspace{-0.04in} \log P\left(y|x; \theta_{base}\right)}{|y|}  \label{q}
\end{eqnarray}
$P\left(y|x; \theta_{base}\right)$ is a baseline model with parameters $\theta_{base}$ trained on the background parallel corpus,  $P\left(y|x; \theta_Z\right)$ is a $Z$-domain model with  $\theta_Z$ by fine-tuning $\theta_{base}$ on a small, $Z$-domain parallel corpus $\widehat{D}_Z$ with trusted quality and $|y|$ is the length of $y$.
 $q_Z$ discerns both noise and domain $Z$ \cite{co-curriculum}. Each domain $Z$ has its own $\widehat{D}_Z$. 


Importantly, \citet{david:2019:internal} shows that, under the Taylor approximation \cite{abramowitz+stegun}, $q_Z$ approximates the dot product between gradient, $g(x,y; \theta_{base})$, of training example $(x, y)$ and  gradient, $g(\widehat{D}_Z, \theta_{base})$, of seed data $\widehat{D}_Z$.\footnote{That is, according to \citet{david:2019:internal}:
\begin{eqnarray}
q_Z(x, y) \times |y| &=& \nonumber \\
\log P(y|x; \theta_Z) - \log P(y|x; \theta_{base}) &\approx & \nonumber \\
\lambda ~ g(x, y; \theta_{base})^\top g(\widehat{D}_Z, \theta_{base}) &&
\label{taylor}
\end{eqnarray}
when  $\theta_{base}$ and $\theta_Z$ are close, which is the case for fine-tuning: $\theta_Z = \theta_{base} + \lambda ~ g(\widehat{D}_Z, \theta_{base})$. 
}
Thus an example with positive $q_Z$ likely moves a model towards domain $Z$. For multiple domains, $Z_1, ..., Z_K$,  selecting a batch of examples with $q_{Z_k}$'s  all being positive would move a model towards a common direction shared across multiple domains, which alleviates forgetting.

The $Z$-domain feature $q_Z\left(x,y\right)$ can be easily generalized into a single {\em multi-domain feature}, $q_{\mathcal{Z}}$, for a set of domains $\mathcal{Z}$:
\begin{eqnarray}
q_{\mathcal{Z}}\left(x,y\right) \hspace{-0.04in} = \hspace{-0.05in}
 \frac{\log P\left(y|x; \theta_{\mathcal{Z}}\right) \hspace{-0.04in} - \hspace{-0.04in} \log P\left(y|x; \theta_{base}\right)}{|y|}  \label{multi_domain_q}
\end{eqnarray}
by simply concatenating all the seed parallel corpus $\widehat{D}_{Z}$ from  the constituent domains into  $\widehat{D}_{\mathcal{Z}}$ and use it to fine-tune the baseline $\theta_{base}$ into $\theta_{\mathcal{Z}}$. 
A benefit of $q_{\mathcal{Z}}$ is scalability: using a single feature value to approximate  $(x, y)$'s gradient consistency with the multiple domains at once. Simple concatenation means, however, domain balancing is not  optimized as in Eq.~\ref{W}.

\paragraph{NLM domain features ($d_Z$)} \cite{Moore2010,dynamiccds} compute $Z$-domain relevance of sentence $x$ with neural language models (NLM),  like $q_Z$:
\begin{eqnarray}
d_Z\left(x\right) = 
   \frac{\log P\left(x; \vartheta_Z\right) - \log P\left(x; \vartheta_{base}\right)}{|x|} \label{varphi}
\end{eqnarray}
where $P(x; \vartheta_{base})$ is an NLM with parameters $\vartheta_{base}$ trained on the $x$ half of the background parallel data, and $P(x; \vartheta_Z)$ is obtained by  fine-tuning $P(x;\vartheta_{base})$ on $Z$-domain monolingual data. Although $d_Z$ may not necessarily reflect the translation gradient of an example under an NMT model, it effectively assesses the $Z$-domain relevance and, furthermore, allows us to include additional larger amounts of in-domain monolingual data. We do not use its bilingual version \cite{Axelrod2011}, but  choose to consider only the source side, for simplicity.

\paragraph{Cross-lingual embedding similarity feature ($\mathrm{emb}$)} computes the cosine similarity of a sentence pair in a cross-lingual embedding space. The embedding model is trained to produce similar representations exclusively for true bilingual sentence pairs, following \citet{Yang2019ImprovingMS}.

\paragraph{BERT quality feature ($\mathrm{BERT}$)} represents quality scores from a fine-tuned multilingual BERT model~\cite{bert}. We fine-tune a pre-trained BERT model\footnote{We use the public cased 12 layers multilingual model: \tt{multi\_cased\_L-12\_H-768\_A-12}} on a supervised dataset with positive and negative translation pairs. 

\vspace{0.1in}
These features compensate each other by capturing the information in a sentence pair from different aspects:
NLM features capture domain. NMT features  additionally discern noise. BERT  and $\mathrm{emb}$ are introduced for denoising, by transfering the strength of the data they are trained on. All these features are from previous research and here we integrate them to solve a generalized problem.

\subsection{Performance Metric $\mathcal{P}$}\label{sec:metric}

Eq.~\ref{W} evaluates the end performance $\mathcal{P}(\widehat{\mathcal{C}}(V))$ of a multi-domain curriculum candidate. We simply combine the validation sets from multi-domains into a single validation set to report the perplexity of the last model checkpoint, after training the model on $\widehat{\mathcal{C}}(V)$. The best multi-domain curriculum minimizes model's perplexity (or maximizes its negative per Eq.~\ref{W}) on the mixed validation set. We experiment with different mixing ratios.


\subsection{Curriculum Optimization}\label{sec:BO}

\begin{algorithm}[t]
\small
  \label{BayesOpt}
  \begin{algorithmic}[1]
   \STATE $\mathcal{H} = \emptyset$; \# Trial history. 
   \STATE $\sigma_0 = \mathrm{GP}$; \# Initialize surrogate model. 
   \STATE $\alpha = \mathrm{EI}$;   \# Initialize acquisition function.
   \STATE $i=1$;
   \WHILE{$i \le T$}
      \STATE $V_i=\arg\max_V \alpha(V; \sigma_{i-1}, \mathcal{H})$; \#
      Predict weight vector $V_i$ by maximizing acquisition function.
      \STATE $p=\mathcal{P}(\widehat{\mathcal{C}}(V_i))$ by fine-tuning NMT on $\widehat{\mathcal{C}}(V_i)$;
      \STATE $\mathcal{H} = \mathcal{H} \cup \{(V_i, p)\}$; \# Update trial history.
      \STATE Estimate $\sigma_i$ with $\mathcal{H}$;
      \STATE $i=i+1$;
   \ENDWHILE
   \STATE return ($V^{\ast}$, $p^{\ast}$) ($\in \mathcal{H}$) with the best performance $p^{\ast}$.
  \end{algorithmic}
 \caption{Bayesian optimization}
  \label{BayesOpt}
\end{algorithm}
We solve Eq.~\ref{W} with Bayesian Optimization (BayesOpt) \cite{Shahriari2016TakingTH} as the optimizer in Figure~\ref{multitask_curriculum}. BayesOpt is derivative-free and can optimize expensive black-box functions, with no 
assumption of the form of $\mathcal{P}$. It has recently become popular for training expensive machine-learning models in the ``AutoML'' paradigm.
It consists of a surrogate model for approximating $\mathcal{P}(\widehat{\mathcal{C}}(V))$ and an acquisition function for deciding the next sample to evaluate. The surrogate model evaluates $\widehat{\mathcal{C}}(V)$ without running the actual NMT training, by the Gaussian process (GP) priors over functions that express assumptions about $\mathcal{P}$.
The acquisition function depends on previous trials, as well as the GP hyper-parameters.
 The Expected Improvement (EI) criterion \cite{Srinivas:2010:GPO:3104322.3104451} is usually used as acquisition function.
 Algorithm~\ref{BayesOpt} depicts how BayesOpt works in our setup.
We use Vizier \cite{vizier} for Batched Gaussian Process Bandit, but open-source implementations of BayesOpt are easily available.\footnote{\url{E.g., https://github.com/tobegit3hub/advisor}}. 






\subsection{Curriculum Construction}\label{curriculum_generation}
We pre-compute all features for each sentence pair $(x,y)$ in training data and turn its features into a single score $f(x,y)$ by Eq.~\ref{varphi2}, given a  weight vector. We then construct a curriculum by instantiating its re-weighting $W_t(x,y)$ (Eq.~\ref{form1}).
To that end, we define a Boolean, dynamic data selection function $\chi_{\rho}^{f}(x, y; t)$ to check, at step $t$, if $(x,y)\in D$ belongs to the top $\rho(t)$-ratio examples in training data $D$ sorted in increasing order of $f(x, y)$, $(0 < \rho \le 1)$. So $\chi_{\rho}^{f}$ is a mask. Suppose $n(t)$ examples are selected by $\chi_{\rho}^{f}(x, y; t)$, the re-weighting will then be
\begin{equation}
W_t(x,y) =  1/n(t) \times \chi_{\rho}^{f}(x, y; t). 
\end{equation}
Filtered examples have zero weights and selected ones are uniformly weighted. We set $\rho(t) = (1/2)^{t/H}$ to decay/tighten over time\footnote{When the training data is small, we can, in practice, let a model warm up before applying the schedule.}, controlled by the hyper-parameter $H$. During training, $\chi_{\rho}^{f}(x, y; t)$ progressively selects higher $f(x,y)$-scoring examples. In implementation, we integrate $\chi_{\rho}^{f}(x, y; t)$ in the data feeder to pass only selected examples to the downstream model trainer; we also normalize $f(x,y)$ offline to directly compare to $\rho(t)$ online to decide filtering.
As an example, the $W_t(x, y)$ for the multi-domain curriculum order in Figure~\ref{curriculum} (5) can look like Table~\ref{curriculum_example} (2).

\section{Experiments}\label{sec:exp}
\subsection{Setup}\label{sec:setup}
\paragraph{Data and domains.} We experiment with two English$\to$French training datasets: the noisy ParaCrawl data\footnote{\url{https://paracrawl.eu}} (290 million sentence pairs) and the WMT14 training data (38 million pairs).
We use SentencePiece model \cite{P18-1007} for subword segmentation with a source-target shared vocabulary of 32,000 subword units. 
We evaluate our method with three ``domains'': two specific domains, news and TED subtitles, and out-of-domain.
News domain uses the WMT14 news testset ({\bf N14}) for testing, and  WMT12-13 for validation in early stopping \cite{Prechelt97automaticearly}.  The TED  domain uses the IWSLT15 testset ({\bf T15}) for testing, and the IWSLT14 testset for validation. Out-of-domain performance is measured by two additional testsets, patent testset ({\bf PA}) (2000 sentences)\footnote{Randomly sampled from \url{www.epo.org}} and WMT15 news discussion testset ({\bf D15}).  We report {\bf SacreBLEU}\footnote{Signature: {\tt BLEU+case.mixed+numrefs.1+smooth .exp+tok.13a+version.1.4.2}}  \cite{clarifybleu}.

\paragraph{Features.} NMT features use the parallel data to train the baseline NMT models. The new-domain-discerning NMT feature $q_N$ uses WMT10-11 (5500 pairs) as in-domain data $\widehat{D}_N$. The TED NMT feature $q_T$ uses the TED subtitle training data (22k pairs) as in-domain data $\widehat{D}_T$.
NLM features use the English half of parallel data to train the baseline NLMs. The news-domain-discerning NLM feature $d_N$ uses the 28 million English sentences from WMT14. The TED subtitle NLM feature $d_T$ uses the English side of IWSLT15 in-domain parallel training data. 
The training of the cross-lingual embedding model follows~\citet{Yang2019ImprovingMS} with a 3-layer transformer~\cite{NIPS2017_7181} (more details in Appendix A.1).
For the BERT feature, we sample positive pairs from the same data to train the cross-lingual embedding model. The negatives are generated using the cross-lingual embedding model, via 10-nearest neighbor retrieval in the embedding space, excluding the true translation. We pick the nearest neighbor to form a hard negative pair with the English sentence, and a random neighbor to form another negative pair.
We sample 600k positive pairs and produce 1.8M pairs in total.


   
\paragraph{Model.} We use LSTM NMT \cite{wu2016} as our models, but with the Adam optimizer \cite{DBLP:journals/corr/KingmaB14}. The batch size is 10k averaged over 8 length-buckets (with synchronous training). NLM/NMT features
 uses 512 dimensions by 3 layers--NLM shares the same architecture as NMT by using
  dummy  source sentences \cite{sennrich-haddow-birch:2016:P16-11}.  The final models are of 1024 dimensions by 8 layers, trained for $55k$ max steps. Training on WMT data uses a dropout probability of 0.2.
Transformer results are in Appendix~\ref{Trans}.


\paragraph{Curriculum optimization.}  In Eq.~\ref{W} (Section~\ref{sec:BO}), we launch $30$ trials (candidate curricula).
 BayesOpt spends $25$ trials in exploration and the last $5$ in exploitation. Each trial trains for $2k$ steps\footnote{
2k is empirically chosen to be practical.  We use a number of fine-tuning trials in Eq.~\ref{W}. NMT training is expensive so we don't want a trial to tune for many steps. NMT is very adaptive on domain data, so each trial does not need many steps. We find no significant difference among 1k, 2k, 6k.
} by fine-tuning a warmed-up model with the candidate curriculum. 
The curriculum decays ($\rho(t)$) from $100\%$ and plateaus at $20\%$ at step 2k. 
We simply and heuristically set a range of $[0.0, 1.0]$ for all feature weights. We don't normalize feature values when weighting them.

\subsection{Results}
We evaluate if the multi-domain curriculum meets requirements (i) and (ii) in Section~\ref{general_idea}.


\subsubsection{Compared to no curriculum}

\begin{table}[t]
\begin{center}
\small
\setlength\tabcolsep{2.8pt} 
\begin{tabular}{|l|cccc|c|}
\hline
 {\bf Curriculum} & 
        {\bf N14 } & 
        {\bf T15 }  &
        {\bf PA} &
        {\bf D15} & {\bf Avg} \\ 
\hline
P1: $\mathcal{B}$ & 33.4 & 35.7 & 29.8 & 30.4 & 32.3 \\
P2: $\mathcal{\widehat{C}}_{6\text{-feats}}$ & {\bf  37.0} & {\bf 38.1} & {\bf 48.3} & {\bf 35.7} & {\bf 39.8} \\
\hline 
W1: $\mathcal{B}$ & 38.0$_{\it 39.2}$ & 37.9 & 45.6 & 34.5  & 39.0 \\
\cite{wu2016} & {\it 39.2} & -- & -- & -- & -- \\
W2: $\mathcal{\widehat{C}}_{6\text{-feats}}$ & {\bf 39.3}  & {\bf 38.8} & {\bf 46.1} & {\bf 36.1} & {\bf 40.1} \\
\hline
\end{tabular}
\end{center}
\vspace{-0.1in}
\caption{\footnotesize
English$\to$French multi-domain curriculum improves no-curriculum baseline ($\mathcal{B}$) over all testsets. Avg: averaged score per row, for ease of reading. P: ParaCrawl data. W: WMT14 training data. \label{BLEU:vs_random}
BLEUs in italics are tokenized BLEU. Other scores are de-tokenized SacreBLEU.
}
\end{table}

We compare:
\begin{itemize}
\setlength\itemsep{0.03pt}
\item $\mathcal{B}$: baseline that does not use curriculum learning. 
\item  $\mathcal{\widehat{C}}_{6\text{-feats}}$: multi-domain curriculum  with 6 features, $d_N$, $d_T$, $q_N$, $q_T$, $\mathrm{BERT}$, $\mathrm{emb}$, weights learned by BayesOpt.
\end{itemize}
Table~\ref{BLEU:vs_random} shows $\mathcal{\widehat{C}}_{6\text{-feats}}$ improves $\mathcal{B}$ on all testsets, especially on noisy ParaCrawl--requirement (i) is met. It is important to note that our WMT baseline (W1) matches \citet{wu2016} on N14, as shown by re-computed tokenized BLEU (italics). 




\subsubsection{Compared to single-domain curricula}

\begin{table}[t]
\begin{center}
\small
\setlength\tabcolsep{2.8pt} 
\begin{tabular}{|l|cccc|c|}
\hline
 {\bf Curriculum} & 
        {\bf N14 } & 
        {\bf T15 }  &
        {\bf PA} &
        {\bf D15} & {\bf Avg} \\ 
\hline
P1: $\mathcal{B}$ & 33.4 & 35.7 & 29.8 & 30.4 & 32.3 \\
\hdashline
P3: $\mathcal{C}_{d_N}$ & 34.7 & 36.2 & 32.6 & 32.6 &  34.0 \\
P4: $\mathcal{C}_{d_T}$ &  34.8 & 36.3 & 30.1 & 32.4 &  33.4 \\
P5.1: $\mathcal{C}_{\mathrm{BERT}}$ & 36.8 & 37.3 & \fbox{47.9} & 35.0 & \fbox{39.3}  \\ 
P5.2: $\mathcal{C}_{\mathrm{emb}}$ & \fbox{36.9} & 37.7 & 46.0 & \fbox{35.2} & 39.0 \\
P6: $\mathcal{C}_{q_N}$ & 36.8 & 37.1 & 47.7 & 34.9 &   39.1 \\ 
P7: $\mathcal{C}_{q_T}$ &  35.6 & \fbox{38.3} &  46.6 & 34.9 & 38.9 \\
\hdashline
P2: $\mathcal{\widehat{C}}_{6\text{-feats}}$ & {\bf  37.0} & 38.1 & {\bf 48.3} & {\bf 35.7} & {\bf 39.8} \\

P2 -- P* & +0.1 & -0.2 & +0.4 & +0.5 & +0.2 \\
\hline 
W1: $\mathcal{B}$ & 38.0 & 37.9 & 45.6 & 34.5  & 39.0 \\
\hdashline
W3: $\mathcal{C}_{d_N}$ & 38.3  & 38.1 & 39.1 & 35.1 &  37.7 \\
W4: $\mathcal{C}_{d_T}$ & 38.1 & 38.4 & 43.0 & \fbox{36.1} &  38.9 \\
W5.1: $\mathcal{C}_{\mathrm{BERT}}$ & \fbox{38.5} & 37.8  & \fbox{45.9} & 35.9 & 39.5 \\ 
W5.2: $\mathcal{C}_{\mathrm{emb}}$ & \fbox{38.5} & 37.8  & 45.8 & 35.9 & 39.5 \\ 

W6: $\mathcal{C}_{q_N}$ & 37.8 & 38.0 & \fbox{45.9} & 35.3 &  39.3 \\ 
W7: $\mathcal{C}_{q_T}$ & \fbox{38.5} & \fbox{38.8} & 45.0  & \fbox{36.1} & \fbox{39.6}  \\
\hdashline
W2: $\mathcal{\widehat{C}}_{6\text{-feats}}$ & {\bf 39.3} & {\bf 38.8}  & {\bf 46.1} & {\bf 36.1} & {\bf 40.1} \\
W2 -- W* & +0.8 & 0.0 & +0.2 & 0.0 & +0.3 \\
\hline
\end{tabular}
\end{center}
\vspace{-0.1in}
\caption{\footnotesize English$\to$French multi-domain curriculum (P2, W2) vs. single-domain curricula (P3-7, W3-7).  Frame boxes mark best per-testset BLEU (W*, P*) over all single-domain curricula. Bold color denotes multi-domain curriculum has best BLEU (W2-W* $\ge$ 0). 
\label{oracle-bleus}
}
\end{table}

We examine the following individual curricula, by training NMT models with each, respectively:
\begin{itemize}
\setlength\itemsep{0.03pt}
    \item $\mathcal{C}_{d_N}$, uses news NLM feature $d_N$ (Eq.~\ref{varphi}).
    \item $\mathcal{C}_{d_T}$, uses TED subtitle NLM feature $d_T$.
    \item $\mathcal{C}_{q_N}$, uses news NMT feature $q_N$ (Eq.~\ref{q}).
    \item $\mathcal{C}_{q_T}$, uses TED  NMT feature $q_T$.
    \item $\mathcal{C}_{\mathrm{BERT}}$, uses BERT quality feature.
    \item $\mathcal{C}_{\mathrm{emb}}$, uses cross-lingual embedding feature.
\end{itemize}
 In Table~\ref{oracle-bleus}, frame boxes  mark the best  BLEUs (P* or W*)  per column, across  P3-P7 or W3-W7. The last column shows averaged BLEU over all testsets.  Bold font indicates $\mathcal{C}_{6\text{-feats}}$ matches or improves W*. As shown, $\mathcal{C}_{6\text{-feats}}$ matches or slightly outperforms the per-domain curricula across testsets. Therefore, $\mathcal{\widehat{C}}_{6\text{-feats}}$ meets requirement (ii).


\subsection{Ablation Studies}
\subsubsection{Features}
\paragraph{Strengths and weaknesses of a feature.} Table~\ref{oracle-bleus} also reveals the relative strengths and weaknesses of each type of features. The peak BLEU (in a frame box) on each testset is achieved by one of $\mathcal{C}_{\mathrm{BERT/emb}}$, $\mathcal{C}_{q_N}$  and $\mathcal{C}_{q_T}$, less by NLM features $d_N$, $d_T$. This contrast seems bigger on the noisy ParaCrawl, but the NLM features do bring gains over $\mathcal{B}$. Overall, $\mathcal{C}_{\mathrm{BERT/emb}}$ (P5, W5) perform well, attributed to their denoising power, but lose to the NMT features (P7, W7) on T15, due to lack of explicit capturing of domain.  The NMT features seem to subtly compensate in domains, and the domain features in denoising, but working with other features improves  the model. 

\begin{table}[t]
\begin{center}
\small
\setlength\tabcolsep{2.8pt} 
\begin{tabular}{|l|cccc|c|}
\hline
 {\bf Curriculum} & 
        {\bf N14 } & 
        {\bf T15 }  &
        {\bf PA} &
        {\bf D15} &{\bf Avg} \\ 
\hline
P2: $\mathcal{\widehat{C}}_{6\text{-feats}}$ & {\bf  37.0} & {\bf 38.1} & {\bf 48.3} & {\bf 35.7} & {\bf 39.8} \\ 
P8: $\mathcal{\widehat{C}}_{4\text{-feats}}$ & 36.6  & {\bf 38.1} & 46.7 & 35.5  &  39.2  \\

\hline 
W2: $\mathcal{\widehat{C}}_{6\text{-feats}}$ &  {\bf 39.3} & 38.8  & 46.1& {\bf 36.1} & {\bf 40.1}  \\
W8: $\mathcal{\widehat{C}}_{4\text{-feats}}$ &   38.9  & {\bf 38.9} &  {\bf 46.5}  & {\bf 36.1} & {\bf 40.1} \\
\hline
\end{tabular}
\end{center}
\vspace{-0.1in}
\caption{\footnotesize BERT and emb features positively contribute to $\mathcal{\widehat{C}}_{6\text{-feats}}$ on ParaCrawl (P).  \label{BLEU:4feats_vs_6feats}}
\end{table}

\paragraph{BERT and emb features.} Both BERT and emb use knowledge external to the experiment setup. For a fair comparison to baselines and a better understanding of them, we drop them by building
\begin{itemize}
\item $\mathcal{\widehat{C}}_{4\text{-feats}}$, multi-domain curriculum that excludes BERT and emb and uses 4 features.
\end{itemize}
 Table~\ref{BLEU:4feats_vs_6feats} shows BERT and emb features in $\mathcal{\widehat{C}}_{6\text{-feats}}$ improve $\mathcal{\widehat{C}}_{4\text{-feats}}$ with ParaCrawl, adding to the intuition that they have a denoising effect.

\begin{figure}[t]
\begin{center}
\begin{tikzpicture}[scale=0.5]
\begin{axis}[ybar,
             xlabel={Feature Name}, ylabel={Feature Value},
             ymax=1.0, ymin=0.0, minor y tick num = 0,
             bar width=7pt, 
             xtick=data, 
             symbolic x coords={$d_\mathrm{N}$, $d_\mathrm{T}$, $q_\mathrm{N}$, $q_\mathrm{T}$, BERT, emb}, 
             legend pos=north west,
             cycle list={{fill=gray!50,draw=black!70},
                  {fill=gray!0,draw=gray!80},
                    {fill=gray!0,draw=gray!0}}
             ],
             
\addplot+[fill,text=black] coordinates { ($d_\mathrm{N}$, 0.00629435 ) ($d_\mathrm{T}$, 0.0252453) ($q_\mathrm{N}$, 0.993615) ($q_\mathrm{T}$, 0.309340) (BERT,0.297420) (emb, 0.965781) };
\addplot+[fill,text=black] coordinates { ($d_\mathrm{N}$, 0.231093) ($d_\mathrm{T}$, 0.440955) ($q_\mathrm{N}$, 0.725233) ($q_\mathrm{T}$, 0.835449) (BERT, 0.513949) (emb, 0.0000925953) } ;
\addlegendentry{Paracrawl}
\addlegendentry{WMT}
\end{axis}
\end{tikzpicture}
\end{center}
\vspace{-0.1in}
\caption{\footnotesize BayesOpt learns to weight features adaptively on ParaCrawl and WMT, respectively.
}
\label{learned_weights}
\end{figure}
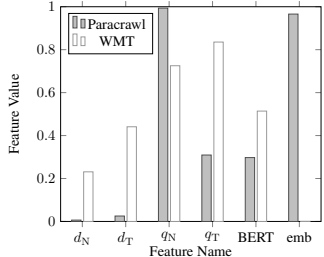

\paragraph{Learned feature weights.} Figure~\ref{learned_weights} shows BayesOpt learns to weight features adaptively in $\mathcal{\widehat{C}}_{6\text{-feats}}$ on ParaCrawl (grey) and WMT (white), respectively. ParaCrawl is very noisy thus noise non-discerning features $d_N$ and $d_T$ do not have a chance to help, but their weights become stronger on the cleaner WMT training data. It is surprising that  BERT feature is still useful to the WMT training. We hypothesize this may suggest  BERT feature  have additional strength to just denoising, or that data noise could be subtle and exist in cleaner data.  


\subsubsection{BayesOpt vs. random search}

We compare BayesOpt (BO) and Random Search (RS) \cite{Bergstra:2012:RSH:2188385.2188395} to solve Eq.~\ref{W}, as well as uniform weighting (Uniform). In Table~\ref{BLEU:bayesopt_vs_rs},  all improve baselines, especially on ParaCrawl (P). RS does  surprisingly well on ParaCrawl, but BayesOpt appears better overall.\footnote{
RS uses 30 trials, as BO (Section~\ref{sec:setup}), so the results show their comparison given the same number of trials.  
}

\begin{table}[t]
\begin{center}
\small
\setlength\tabcolsep{2.8pt} 
\begin{tabular}{|l|cccc|c|}
\hline
 {\bf Curriculum} & 
        {\bf N14 } & 
        {\bf T15 }  &
        {\bf PA} &
        {\bf D15} & {\bf Avg} \\ 
\hline
P1 : $\mathcal{B}$ &  33.4 & 35.7 & 29.8 & 30.4 & 32.3 \\
P2 : $\mathcal{\widehat{C}}_{6\text{-feats}}$ (BO)   & {\bf  37.0} & 38.1 & {\bf 48.3} & {\bf 35.7} & {\bf 39.8} \\
P9 : $\mathcal{\widehat{C}}_{6\text{-feats}}$ (RS)  & 36.7 & {\bf 38.4} & 48.0 & 35.5 & 39.7 \\
P10: $\mathcal{\widehat{C}}_{6\text{-feats}}$ (Uniform)  & 35.4 & 36.9 & {\bf 48.3} & 34.1 & 38.7 \\
\hline 
W1 : $\mathcal{B}$ & 38.0 & 37.9 & 45.6 & 34.5  & 39.0 \\
W2 : $\mathcal{\widehat{C}}_{6\text{-feats}}$ (BO)  & {\bf 39.3} & 38.8  & {\bf 46.1} & 36.1 & {\bf 40.1} \\
W9 : $\mathcal{\widehat{C}}_{6\text{-feats}}$ (RS) &  39.0  & 38.2 & 43.7 & {\bf 36.4} & 39.3\\
W10: $\mathcal{\widehat{C}}_{6\text{-feats}}$ (Uniform) &  38.8  & {\bf 39.1} & 43.0 & 36.0 & 39.2\\
\hline
\end{tabular}
\end{center}
\vspace{-0.1in}
\caption{\footnotesize On average, BayesOpt (BO) performs better than Random Search (RS) and uniform weighting (Uniform), for learning feature weights of a multi-domain curriculum. 
\label{BLEU:bayesopt_vs_rs}}
\end{table}

\subsubsection{Mixing validation sets}
Eq.~\ref{W} evaluates $\mathcal{P}$ using the concatenated validation set (Section~\ref{sec:metric}).
Table~\ref{mixdev} shows that the news-vs-TED mixing ratios can affect the per-domain BLEUs. For example, on ParaCrawl, when news sentences are absent from the validation set, N14 drops by 0.7 BLEU (P8 vs. P13). We use the four feats as in $\mathcal{\widehat{C}}_{4\text{-feats}}$ in this examination.

\begin{figure}[t]
\begin{center}
\begin{tikzpicture}[scale=0.5]
\begin{axis}[xlabel={Selection Threshed = 1-$\rho(t)$ (Section~\ref{curriculum_generation})},
   ylabel={Domain Relevance}, ybar stacked, bar width=5,
yticklabel style={%
        /pgf/number format/.cd,
            fixed,
            fixed zerofill,
            precision=2,
    },
    axis y line*=left,
    scale only axis,
    legend pos = south east,
    cycle list={
  {fill=gray!50,draw=black!70},
  {fill=gray!0,draw=gray!80},
  {fill=gray!0,draw=gray!0}
  }
]

\addplot table[x index=0,y index=1,col sep=comma] {curve.txt};
\addlegendentry{news}
\label{pgfplots:plot2}

\addplot table[x index=0,y index=2,col sep=comma] {curve.txt};
\addlegendentry{TED subtitles}
\label{pgfplots:plot3}

\end{axis}

\begin{axis}[scale only axis, xlabel={Selection Threshed = 1-$\rho(t)$ (Section~\ref{curriculum_generation})},ylabel={Quality},  bar width=5,
yticklabel style={%
        /pgf/number format/.cd,
            fixed,
            fixed zerofill,
            precision=1,
    },
    axis y line*=right,
    axis x line=none,
    ylabel near ticks,
    legend pos = north west,
    cycle list name=black white,
  ]

\addplot table[x index=0,y index=3,col sep=comma] {curve.txt};

\addlegendentry{human rating}
\label{pgfplots:plot1}
\end{axis}

\end{tikzpicture}
\end{center}
\vspace{-0.1in}
\caption{\footnotesize The multi-domain curriculum dynamically balances multi-domain-relevant and noise-reduced data, as validated by human ratings.
}
\label{dynamic}
\end{figure}
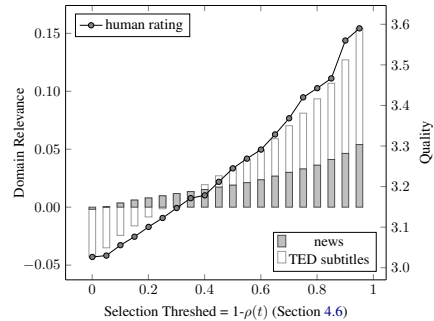

\begin{table}[t]
\setlength\tabcolsep{2.8pt} 
\begin{center}
\small
\begin{tabular}{|lc|cccc|c|}
\hline
&{\bf Mixing Ratio} & {\bf N14 }  & {\bf  T15}  & {\bf PA} & {\bf D15} & {\bf Avg} \\ 
\hline 
P11:& 1.0:0.0 & 36.3 & 37.8 & 47.3 & 35.3 &  39.2 \\
P12:& 0.8:0.2 & 36.4 & {\bf 38.2} & {\bf 47.7} & 35.4 &  {\bf 39.4} \\
P8:  &0.5:0.5 & {\bf 36.6}  & 38.1 & 46.7 & {\bf 35.5}  &  39.2 \\
P13: & 0.0:1.0 & 35.9 & 38.1 & 47.0 & 35.2 &  39.1 \\
\hline
W11:& 1.0:0.0 & 39.1  & 38.6  & 46.4 & 36.0 & 40.0 \\
W12:& 0.8:0.2 &  39.0 & 38.7 & 46.3 & 35.7 & 39.9 \\
W8:  & 0.5:0.5 & 38.9  & {\bf 38.9} &  {\bf 46.5}  & {\bf 36.1} & {\bf 40.1} \\
W13: & 0.0:1.0 & {\bf 39.1} & 38.6 & 46.4 & 36.0 & 40.0  \\
\hline 
\end{tabular}
\end{center}
\vspace{-0.1in}
\caption{\small Guiding multi-domain curriculum learning by mixing validation sets. Experiments use 4 features as in $\mathcal{\widehat{C}}_{4\text{-feats}}$.
\label{mixdev}
}
\end{table}

\subsubsection{Dynamic data balancing} \label{dynamic_sampling}

We simulate dynamic data selection with a random sample of 2000 pairs from the WMT data and annotate each pair by human raters with 0 (nonsense) - 4 (perfect) quality scale (following \citet{denoise_nmt}). We sort the pairs by $f(x, y)$ (Eq.~\ref{varphi2}). A threshold selects a subset of pairs, for which we average the respective NMT feature values as the domain relevance. Figure~\ref{dynamic} shows that the multi-domain curriculum ($\mathcal{\widehat{C}}_{6\text{-feats}}$) learns to  dynamically increase quality and multi-domain relevance. 
Therefore, our idea (Section~\ref{general_idea}) works as intended. Furthermore, training seems to gradually increase quality or domain in different speeds, determined  by Eq.~\ref{W}.

\subsubsection{Weighting loss vs. curriculum}
With the learned weights, we compute a weight for each example to sort data to form a curriculum. Alternatively, we could weight the cross-entropy loss for that sentence during training \citep{D17-1155,chen-etal-2017-cost}. Table~\ref{weighted-loss} shows that curriculum yields improvements over weighing per-sentence loss, in particular on noisy training data, confirming previous findings \citep{dynamiccds}.

\begin{table}[t]
\small
\begin{center}
\begin{tabular}{|l|cccc|c|}
\hline
{\bf Model} & {\bf N14 }  & {\bf  T15}  & {\bf PA} & {\bf D15} & {\bf Avg} \\ 
\hline
P8: \hspace{0.03in} Curriculum  & {\bf 36.6} & {\bf 38.1} & {\bf 46.7} & {\bf 35.5} & {\bf 39.2} \\
P14: Weight Loss  & 35.3 & 37.8 & 39.3 & 32.6 & 36.3 \\
\hline
W8: \hspace{0.03in} Curriculum & {\bf 38.9} & {\bf 38.9} & {\bf 46.5} & {\bf 36.1} & {\bf 40.1} \\
W14: Weight Loss  & 38.6  & 37.6 & 45.7 & 35.3 & 39.3 \\
\hline
\end{tabular}
\end{center}
\caption{\small Forming a curriculum with learned weights performs better than
weighting instance loss in training. Experiments use 4 features (as in $\mathcal{\widehat{C}}_{4\text{-feats}}$).
\label{weighted-loss}
}
\end{table}

\subsubsection{In-domain fine-tuning}

\begin{table}[t]
\begin{center}
\small
\setlength\tabcolsep{2.8pt} 
\begin{tabular}{|l|cccc|c|}
\hline
{\bf Model } & {\bf N14 }  & {\bf  T15}  & {\bf PA} & {\bf D15} & {\bf Avg} \\ 
\hline
P15: $\mathcal{B}$+N & 35.8 & 37.1 & 41.2 & 32.8 & 36.7  \\
P16: $\mathcal{B}$+T  &  35.8 &  38.7 & 45.4 & 34.6 &  38.6 \\
P17: $\mathcal{B}$+N+T  &  35.9 &  38.7 & 44.8 & 34.4 &  38.4 \\
\hdashline
P2\hfill: $\mathcal{\widehat{C}}_{6\text{-feats}}$ (BO)   & 37.0 & 38.1 & 48.3 & 35.7 & 39.8 \\

P18: $\mathcal{{\widehat{C}}}_{6\text{-feats}}$ +N+T & {\bf 38.1} & {\bf 39.7} & {\bf 48.6} & {\bf 36.6} & {\bf 40.8} \\
\hline
W15: $\mathcal{B}$+N & 38.7 & 37.4 & {\bf 46.4}  & 34.6 &  39.3 \\
W16: $\mathcal{B}$+T  & 36.8& 38.9& 44.8 & 36.5 & 39.3 \\
W17: $\mathcal{B}$+N+T  & 38.6& 39.1& 46.1 & 35.8 & 39.9 \\
\hdashline
W2\hfill: $\mathcal{\widehat{C}}_{6\text{-feats}}$ (BO)  & {\bf 39.3} & 38.8  & 46.1 & 36.1 & 40.1 \\
W18: $\mathcal{\widehat{C}}_{6\text{-feats}}$ +N+T  & {\bf 39.3}  & {\bf 39.8} & 46.0 &  {\bf 36.6} & {\bf 40.4 } \\
\hline
\end{tabular}
\end{center}
\caption{\footnotesize The multi-domain curricula still bring improvements, even after models are fine-tuned on in-domain parallel data.
+N: fine-tune on news parallel data $\widehat{D}_N$ (Section~\ref{sec:setup});
+T: fine-tune on TED parallel data $\widehat{D}_T$;
+N+T on concatenation.
\label{ft}
}
\end{table}

$\mathcal{C}_{q_N}$ and $\mathcal{C}_{q_T}$ each use a small in-domain parallel dataset, but we can simply fine-tune the final models on either dataset (+N, +T) or their concatenation (+N+T). Table~\ref{ft} shows that $\widehat{\mathcal{C}}_{\text{6-feats}}$ can be further improved by in-domain fine-tuning\footnote{We fine-tune with SGD for 20k steps, with batch size 16, learning rate 0.0001. } and that both $\widehat{\mathcal{C}}_{\text{6-feats}}$ and its fine-tuning still improve the fine-tuned baselines, in particular on ParaCrawl. 

\subsection{Discussion: Feature Dependency}
One potential issue with using multiple {\em per-domain} features ($q_{{Z}}(x,y)$'s in Eq.~\ref{q}) is  scores are not shared across domains and linear weighting may not capture feature dependency. For example, we need two NMT features if there are two domains. We replace the two NMT features, $q_{{N}}$ and $q_{{T}}$, in $\mathcal{\widehat{C}_\text{4-feats}}$ with a single two-domain feature  $q_{\mathcal{\mathcal{Z}}=\{N, T\}}$ (Eq.~\ref{multi_domain_q}), but with the two corresponding NLM features unchanged (so the new experiment has 3 features).
Table~\ref{multitask-scoring} shows multi-domain feature contributes slightly better than linear combination of  per-domain features (P19 vs. P8).  The per-domain features, however, have the advantage of efficient feature weighting. In case of many features, learning to compress them
seems to be an interesting future investigation.

\begin{table}[t]
\small
\begin{center}
\begin{tabular}{|l|cccc|c|}
\hline
{\bf Model} & {\bf N14 }  & {\bf  T15}  & {\bf PA} & {\bf D15} & {\bf Avg} \\ 
\hline 
P8: per-domain  & {\bf 36.6} & 38.1 & 46.7 & 35.5 & 39.2 \\
P19: multi-domain  & {\bf 36.6}   & {\bf 38.6} & {\bf 46.8} & {\bf 35.9} & {\bf 39.5} \\
\hline
\end{tabular}
\end{center}
\caption{\small Multi-domain/task feature (Eq.~\ref{multi_domain_q}) seems to contribute slightly better than linear combination of  multiple per-domain features (Eq.~\ref{q}).
\label{multitask-scoring}
}
\end{table}

\section{Conclusion}
Existing curriculum learning research in NMT focuses on a single domain. We present a multi-domain curriculum learning method. We carefully introduce instance-level features and learn a training curriculum to 
gradually concentrate on multi-domain relevant and noise-reduced data batches. 
End-to-end experiments and ablation studies on large datasets at
different noise levels show that the multi-domain curriculum simultaneously reaches or outperforms
the individual performance and brings solid gains over no-curriculum training, on in-domain and out-of-domain testsets.

\section*{Acknowledgments}
The authors would like to thank David Grangier for Eq.~\ref{taylor} and derivation, the three anonymous reviewers for their insightful reviews,
Yuan Cao for his technical suggestions, Jason Smith, Markus Freitag, Pidong Wang and Reid Pryzant for comments on an earlier draft,
Quoc V. Le for suggestions in a related thread.
\newpage

\bibliography{acl2020}
\bibliographystyle{acl_natbib}

\appendix
\section{Appendices}
\label{sec:appendix}

\subsection{Cross-lingual Embedding Model Parameters}
The sentence encoder has a shared 200k token multilingual vocabulary with 10k OOV buckets. For each token, we also extract character n-grams ($n=[3, 6]$) hashed to 200k buckets. Word token items and character n-gram items are mapped to $320$ dim.\ character embeddings. Word and character n-gram representations are summed together to produce the final input token representation.  The encoder is a 3-layer Transformer with hidden size of $512$, filter size of $2048$, and $8$ attention heads.
We train for 40M steps using an SGD optimizer with batch size K=100 and learning rate $0.003$. 
During training, the word and character embeddings are scaled by a gradient multiplier of 25. 

\subsection{Transformer-Big Results}
\label{Trans}
We replicate experiments with the Transformer-Big architecture. Table~\ref{Trans_BLEU:vs_random} shows the  Transformer-Big results that correspond to the RNN results in Table~\ref{BLEU:vs_random}. These results show that the multi-domain curriculum  meets the performance requirement (i) (Section~\ref{general_idea}) using the Transformer architecture. Table~\ref{Trans_oracle-bleus} shows the Transformer-Big results corresponding to RNN results in Table~\ref{oracle-bleus}. They show that the proposed multi-domain curriculum meets the performance requirement (ii) using Transformer.
\begin{table}[h]
\begin{center}
\small
\setlength\tabcolsep{2.8pt} 
\begin{tabular}{|l|cccc|c|}
\hline
 {\bf Curri.} & 
        {\bf N14 } & 
        {\bf T15 }  &
        {\bf PA} &
        {\bf D15} & {\bf Avg} \\ 
\hline
P1: $\mathcal{B}$ & 34.1 & 36.3 & 34.2 & 32.3 & 34.2 \\
P2: $\mathcal{\widehat{C}}_{6\text{-feats}}$ & {\bf 39.6} & {\bf 40.2} & {\bf 50.6} & {\bf 37.7} & {\bf 42.0} \\
\hline 
W1: $\mathcal{B}$ & 40.8 & 39.9 & 46.0 & 37.8 & 41.1 \\
W2: $\mathcal{\widehat{C}}_{6\text{-feats}}$ & {\bf 41.8 } & {\bf 41.2 } & {\bf 48.1} & {\bf 38.8} & {\bf 42.5} \\
\hline
\end{tabular}
\end{center}
\caption{Transformer Big SacreBLEU: English $\to$ French multi-domain curriculum improves no-curriculum baseline ($\mathcal{B}$) over all testsets, using Transformer-Big.  P: Paracrawl training data. W: WMT14  training data. \label{Trans_BLEU:vs_random}}
\end{table}

\begin{table}[h]
\begin{center}
\small
\setlength\tabcolsep{2.8pt} 
\begin{tabular}{|l|cccc|c|}
\hline
 {\bf Curri.} & 
        {\bf N14 } & 
        {\bf T15 }  &
        {\bf PA} &
        {\bf D15} & {\bf Avg} \\ 
\hline
P1: $\mathcal{B}$ & 34.1 & 36.3 & 34.2 & 32.3 & 34.2 \\
\hdashline
P3: $\mathcal{C}_{d_N}$ & 33.7 & 36.1 & 32.7 & 32.5 & 33.8 \\
P4: $\mathcal{C}_{d_T}$ &  35.3 & 37.7 & 32.8 & 34.0 & 35.0 \\
P5: $\mathcal{C}_{\mathrm{BERT}}$ & \fbox{39.2} & 40.1 & \fbox{49.7} & \fbox{37.5} & \fbox{41.6} \\
P6: $\mathcal{C}_{q_N}$ & 38.9 & 39.8 & 48.9 & 36.9 & 41.1  \\ 
P7: $\mathcal{C}_{q_T}$ & 37.3 & \fbox{40.4} &  44.7 & 36.2 & 39.7 \\
\hdashline
P2: $\mathcal{\widehat{C}}_{6\text{-feats}}$ & {\bf  39.6} & 40.2 & {\bf 50.6} &  {\bf 37.7 } &  {\bf 42.0} \\

P2 -- P* & +0.4 & -0.2 & +0.9 & +0.2 & +0.3 \\
\hline 
W1: $\mathcal{B}$ & 40.8 & 39.9 & 46.0 & 37.8 & 41.1 \\
\hdashline
W3: $\mathcal{C}_{d_N}$ & 40.9  & 39.2 & 44.4 & 37.6 & 40.5 \\
W4: $\mathcal{C}_{d_T}$ & 39.8  & 39.6 & 43.3 & 37.3 & 40.0 \\
W5: $\mathcal{C}_{\mathrm{BERT}}$ & 40.5 & 39.2  & 45.7 & 38.3 & 40.9 \\ 

W6: $\mathcal{C}_{q_N}$ & \fbox{41.1} & 40.0 & 47.6 & 38.0 & 41.7 \\ 
W7: $\mathcal{C}_{q_T}$ & \fbox{41.1} & \fbox{41.4} & \fbox{47.7} & \fbox{38.5} & \fbox{42.2}  \\
\hdashline
W2: $\mathcal{\widehat{C}}_{6\text{-feats}}$ & {\bf 41.8}  & 41.2 & {\bf 48.1} & {\bf 38.8} & {\bf 42.5} \\
W2 -- W* & +0.7 & -0.2 & +0.4 & +0.3 & +0.3 \\
\hline
\end{tabular}
\end{center}
\caption{Transformer Big SacreBLEU: English $\to$ French multi-domain curriculum (P2, W2) vs. single-domain curricula (P3-7, W3-7). BLEU scores over 4 testsets and their average.  Frame boxes mark best per-testset BLEU (W*, P*) over all single-domain curricula. Bold color denotes multi-domain curriculum has best BLEU (W2-W* $\ge$ 0). P: ParaCrawl training data. W: WMT14 training data.
\label{Trans_oracle-bleus}
}
\end{table}

\subsection{An Explanation: Noisy Data Useful in Low-Resource Setup}
\label{noisy_exp}
With noisy, limited data (e.g., 100k pairs), we can train a model A on all data, or a model B on the filtered subset (e.g., 10k). We can also fine-tune A on the filtered data, to produce model C. C could be better than A due to use of higher-quality data or better than B due to use of more data (200k>10k). Therefore, by ''noisy data can be helpful'', we refer to data regularization (using more data) and to transfer learning (fine-tuning) to exploit both data quantity and quality, the idea behind dynamic data selection.

\end{document}